\documentclass[10pt,twocolumn,letterpaper]{article}

\usepackage{iccv}
\usepackage{times}
\usepackage{epsfig}
\usepackage{graphicx}
\usepackage{amsmath}
\usepackage{amssymb}

\usepackage{subfigure}
\usepackage{booktabs} 

\usepackage{multirow}
\usepackage[utf8]{inputenc} 
\usepackage[T1]{fontenc}    

\usepackage{url}            
\usepackage{amsfonts}       
\usepackage{nicefrac}       
\usepackage{microtype}      

\usepackage[pagebackref=true,breaklinks=true,letterpaper=true,colorlinks,bookmarks=false]{hyperref}

\iccvfinalcopy 

\def\iccvPaperID{3207} 

\ificcvfinal\fi

\begin{document}

\title{SVGA-Net: Sparse Voxel-Graph Attention Network for 3D Object Detection from Point Clouds}

\author{Qingdong He,
    Zhengning Wang\thanks{Corresponding Author},
    Hao Zeng,
    Yi Zeng,
    Yijun Liu\\
University of Electronic Science and Technology of China\\
{\tt\small heqingdong@alu.uestc.edu.cn,zhengning.wang@uestc.edu.cn} \\
{\tt\small \{haozeng,zengyii,yijunliu\}@std.uestc.edu.cn}
}

\maketitle
\ificcvfinal\thispagestyle{empty}\fi

\begin{abstract}
   Accurate 3D object detection from point clouds has become a crucial component in autonomous driving. However, the volumetric representations and the projection methods in previous works fail to establish the relationships between the local point sets. In this paper, we propose Sparse Voxel-Graph Attention Network (SVGA-Net), a novel end-to-end trainable network which mainly contains voxel-graph module and sparse-to-dense regression module to achieve comparable 3D detection tasks from raw LIDAR data. Specifically, SVGA-Net constructs the local complete graph within each divided 3D spherical voxel and global KNN graph through all voxels. The local and global graphs serve as the attention mechanism to enhance the extracted features. In addition, the novel sparse-to-dense regression module enhances the 3D box estimation accuracy through feature maps aggregation at different levels. Experiments on KITTI detection benchmark demonstrate the efficiency of extending the graph representation to 3D object detection and the proposed SVGA-Net can achieve decent detection accuracy.
\end{abstract}

\section{Introduction}

With the widespread popularity of LIDAR sensors in autonomous driving ~\cite{geiger2012we} and augmented reality~\cite{Park:2008:MOT:1605298.1605357}, 3D object detection from point clouds has become a mainstream research direction. Compared to RGB images from video cameras, point clouds could provide accurate depth and geometric information ~\cite{yu2020point} which can be used not only to locate the object, but also to describe the shape of the object ~\cite{zhang2020monocular}. However, the properties of unordered, sparsity and relevance of point clouds make it a challenging task to utilize point clouds for 3D object detection directly.

In recent years, several pioneering approaches have been proposed to tackle these challenges for 3D object detection on point clouds. The main ideas for processing point clouds data are to project point clouds to different views\cite{simon2019complexer,chen2017multi,ku2018joint,liang2018deep,yang2018pixor} or divide the point clouds into equally spaced voxels\cite{li20173d,zhou2018voxelnet,yan2018second}. Then convolutional neural networks and mature 2D objection detection frameworks~\cite{ren2015faster,redmon2016you} are applied to extract features. However, because projection alone cannot capture the object's geometric information well, many methods\cite{chen2017multi,wang2019frustum,qi2018frustum,8794195} have to combine RGB images in the designed network. While the methods using only voxelization do not make good use of the properties of the point clouds and bring a huge computational burden\cite{liu2019point} as resolution increases. Apart from converting point clouds into other formats, some works~\cite{shi2019pointrcnn,yang2019std} take Pointnets~\cite{qi2017pointnet,qi2017pointnet++} as backbone to process point clouds directly. Although Pointnets build a hierarchical network and use a symmetric function to maintain permutation invariance, they fail to construct the neighbour relationships between the grouped point sets~\cite{Wang:2019:DGC:3341165.3326362}.

Considering the properties of point clouds, we should notice the superiority of graphs in dealing with the irregular data. In fact, in the domain of point clouds for segmentation and classification tasks, the method of processing with graphs has been deeply studied by many works~\cite{8237818,9010397,Landrieu2018Large,8578576,Wang:2019:DGC:3341165.3326362}. However, few researches have used graphs to make 3D object detection from point clouds. To our knowledge, Point-GNN\cite{Point-GNN} may be the first to prove the potential of using the graph neural network as a new approach for 3D object detection. Point-GNN introduces auto-registration mechanism to reduce translation variance and designs box merging and scoring operation to combine detection results from multiple vertices accurately. However, similar to ShapeContextNet~\cite{8578582} and Pointnet++~\cite{qi2017pointnet++}, the relationship between point sets is not well established in the feature extraction process and a large number of matrix operations will bring heavy calculation burden and memory cost.

In this paper, we propose the sparse voxel-graph attention network (SVGA-Net) for 3D object detection. SVGA-Net is an end-to-end trainable network which takes raw point clouds as input as outputs the category and bounding boxes information of the object. Specifically, SVGA-Net mainly consists of voxel-graph network module and sparse-to-dense regression module. Instead of normalized rectangle voxels, we divide the point clouds into 3D spherical space with a fixed radius. The voxel-graph network aims to construct local complete graph for each voxel and global KNN graph for all voxels. The local and global serve as the attention mechanism that can provide a parameter supervision factor for the feature vector of each point. In this way, the local aggregated features can be combined with the global point-wise features. Then we design the sparse-to-dense regression module to predict the category and 3D bounding box by processing the features at different scales. Evaluations on KITTI benchmark demonstrate that our proposed method can achieve comparable results with the state-of-the-art approaches.

Our key contributions can be summarized as follows:
\begin{itemize}
\item We propose a new end-to-end trainable 3D object detection network from point clouds which uses graph representations without converting to other formats.
\item We design a voxel-graph network, which constructs the local complete graph within each spherical voxel and the global KNN graph through all voxels to learn the discriminative feature representation simultaneously.
\item We propose a novel 3D boxes estimation method that aggregates features at different scales to achieve higher 3D localization accuracy.
\item Our proposed SVGA-Net achieves decent experimental results with the state-of-the-art methods on the challenging KITTI 3D detection dataset.
\end{itemize}

\section{Related Work}
\subsection{Projection-based methods for point clouds}
To align with RGB images, series of works process point clouds through projection ~\cite{chen2017multi,ku2018joint,liang2019multi}. Among them, MV3D~\cite{chen2017multi} projects point clouds to bird view and trains a Region Proposal Network (RPN) to generate positive proposals. It extracts features from LiDAR bird view, LIDAR front view and RGB image, for every proposal to generate refined 3D bounding boxes. AVOD~\cite{ku2018joint} improves MV3D by fusing image and bird view features and merges features from multiple views in the RPN phase to generate positive proposals. Note that accurate geometric information may be lost in the high-level layers with this scheme.
\subsection{Volumetric methods for point clouds}
Another typical method for processing point clouds is voxelization. VoxelNet~\cite{zhou2018voxelnet} is the first network to process point clouds with voxelization, which use stacked VFE layers to extract features tensors. Following it, a large number of methods~\cite{liu2019tanet,yan2018second,shi2020pv,chen2019fast} divide the 3D space into regular grids and group the points in a grid as a whole. However, they often need to stack heavy 3D CNN layers to realize geometric pose inference which bring large computation.
\subsection{Pointnet-based methods for point clouds}
To process point clouds directly, PointNet~\cite{qi2017pointnet} and PonintNet++~\cite{qi2017pointnet++} are the two groundbreaking works to design parallel MLPs to extract features from the raw irregular data, which improve the accuracy greatly. Taking them as backbone, many works~\cite{shi2019pointrcnn,qi2018frustum,lang2019pointpillars,yang2019std,yang20203dssd} begin to design different feature extractors to achieve better performance. Although Pointnets are effective to abstract features, they still suffer feature loss between the local and global point sets.
\subsection{Graph-based methods for point clouds}
Constructing graphs to learn the order-invariant representation of the irregular point clouds data has been explored in classification and segmentation tasks~\cite{kaul2019sawnet,Wang:2019:DGC:3341165.3326362}. Graph convolution operation is efficient to compute features between points. DGCNN~\cite{Wang:2019:DGC:3341165.3326362} proposes EdgeConv in the neighbor point sets to fuse local features in a KNN graph. SAWNet~\cite{kaul2019sawnet}  extends the ideas of PointNet and DGCNN to learn both local and global information for points. Surprisingly, few researches have considered applying graph for 3D object detection. Point-GNN may be the first work to design a GNN for 3D object detection. Point-GNN~\cite{Point-GNN} designs a one-stage graph neural network to predict the category and shape of the object with an auto-registration mechanism, merging and scoring operation, which demonstrate the potential of using the graph neural network as a new approach for 3D object detection.

\section{Proposed Method}
\begin{figure*}[htb]
  \centering
  \includegraphics[scale=0.54]{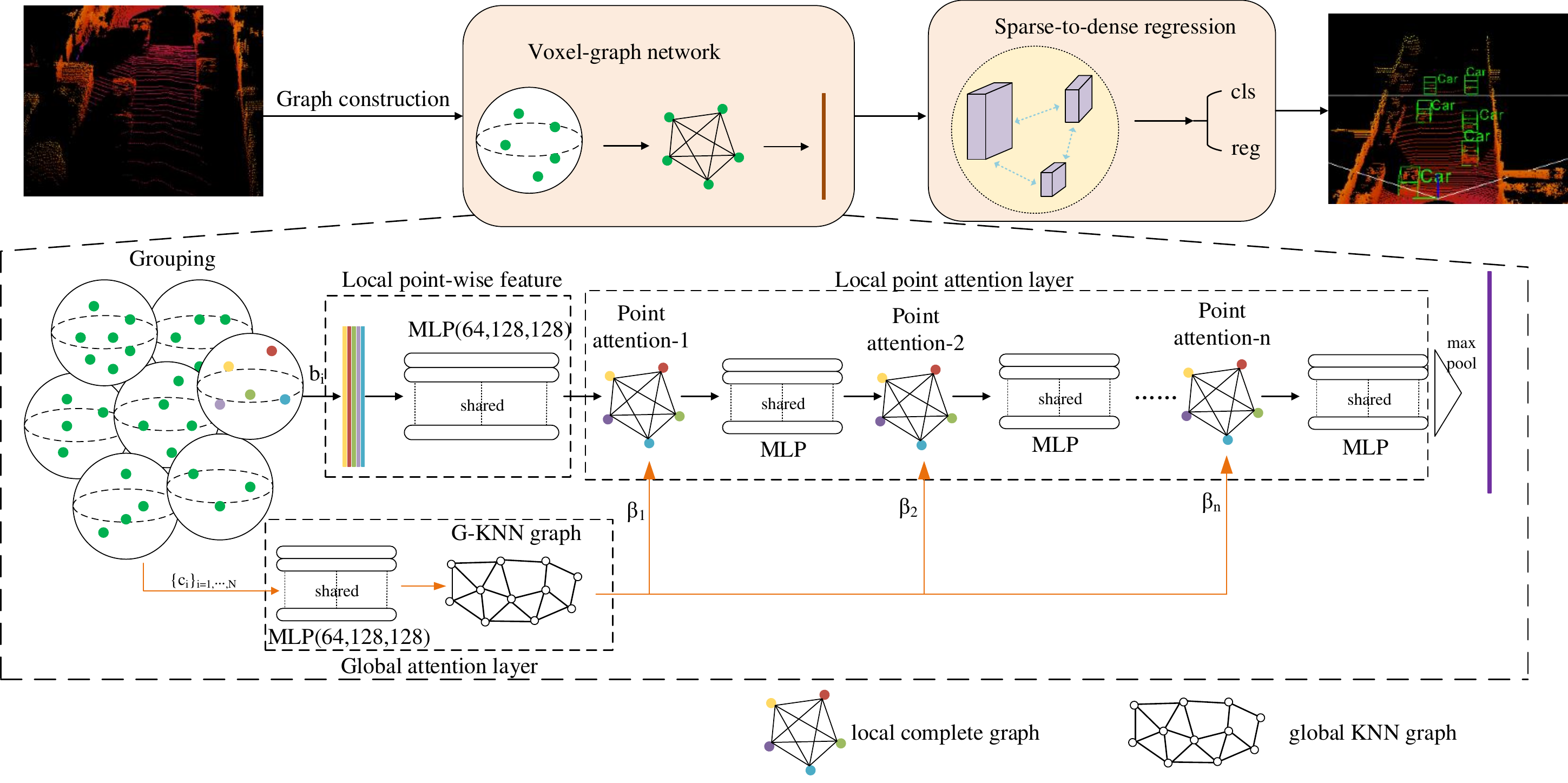}
  \caption{Architecture of the proposed SVGA-Net. The voxel-graph network takes raw point clouds as input, partitions the space into spherical voxels, transforms the points in each sphere to a vector representing the feature information. The sparse-to-dense regression module takes the aggregated features as input as generates the final boxes information.}
  \label{fig:1}
\end{figure*}

In this section, we detail the architecture of the proposed SVGA-Net for 3D detection from point clouds. As shown in Figure~\ref{fig:1}, our SVGA-Net architecture mainly consists of two modules: voxel-graph network and spare-to-dense regression.

\subsection{Voxel-graph network architecture}
\textbf{Spherical voxel grouping.} Consider the original point clouds are represented as $G=\{V, D\}$, where $V = \{p_1,p_2,...,p_n\}$ indicting $n$ points in a $D$ dimensional metric space. In our practice, $D$ is set to 4 so each point in 3D space is defined as $v_i=[x_i, y_i, z_i]$, where $x_i,y_i,z_i$ denote the coordinate values of each point along the axes X, Y, Z and the fourth dimension is the laser reflection intensity which denoted as $s_i$.

Then in order to cover the entire point set better, we use the iterative farthest point sampling~\cite{qi2017pointnet++} to choose $N$ farthest points $P=\{p_i=[ v_i, s_i]^T \in {R}^4\}_{i=1,2,...N}$. According to each point in $P$, we search its nearest neighbor within a fixed radius $r$ to form a local voxel sphere:
\begin{equation}
\label{eqn:01}
b_i=\{p_1, p_2, ... p_i, ..., p_j, ...\mid \parallel v_i - v_j\parallel_2 < r\}
\end{equation}
In this way, we can subdivide the 3D space into $N$ 3D spherical voxels $B=\{b_1, b_2, ..., b_N\}$.

\textbf{Local point-wise feature.} As shown in Figure~\ref{fig:1}, for each spherical voxel $b_i=\{p_j=[x_j, y_j, z_j, s_j]^T\}_{j=1,2,...,t}$ with $t$ points ($t$ varies for different voxel sphere), the coordinate information of all points inside form the input vector. We extract the local point-wise features for each voxel sphere by learning a mapping:
\begin{equation}
\label{eqn:01}
f(b_i) = MLP(p_j)_{j=1,2,...,t}
\end{equation}
Then, we could obtain the local point-wise feature representation for each voxel sphere $F=\{f_i, i=1,...,t\}$, which are transformed by the subsequent layers for deeper feature learning.

\begin{figure}[htb]
  \centering
  \includegraphics[width=0.8\linewidth]{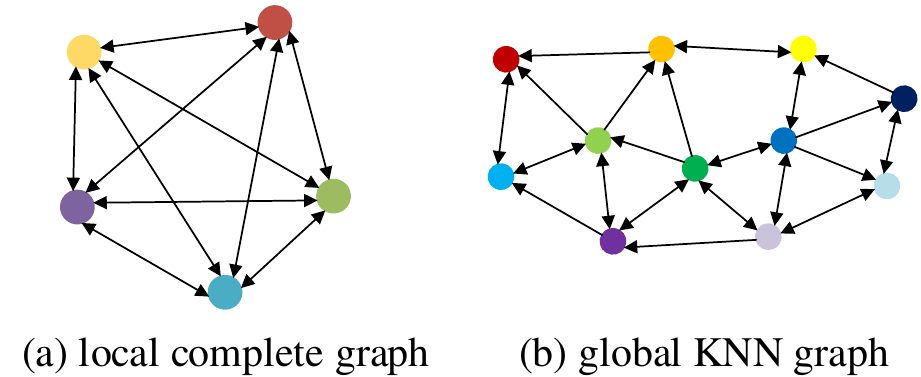}
  \caption{Graph construction. Each node with different color indicates the aggregated feature and arrows direction represents the information propagation direction with independent attention calculations scores. (a) local complete graph: for each node, we aggregate the information of all the nodes within the same spherical voxel according to the attention score. (b) global 3-NN graph: we aggregate the information of the three nearest neighbours around each node according to the attention score.}
  \label{fig:2}
\end{figure}
\textbf{Local point-attention layer.} Taken the features of each nodes as input, the local point-attention layer outputs the refined features $F^{'}=\{f^{'}_i,i=1,...,t\}$ through series of information aggregation. As shown in Figure~\ref{fig:2}, we construct a complete graph for each local node set and KNN graph for all the spherical voxels. We aggregate the information of each node according to the local and global attention score. The feature aggregation of $j$-th node is represented as:
\begin{equation}
\label{eqn:01}
f^{'}_j = \beta_m \cdot f_j + \sum_{k\in \sqcup(p_j)} \alpha_{j,k}\cdot f_{j,k}
\end{equation}
where $f^{'}_j$ denotes the dynamic updated feature of node $p_j$ and $f_j$ is the input feature of node $p_j$. $\sqcup(p_j)$ denotes the index of the other nodes inside the same sphere. $f_{j,k}$ denotes the feature of the $k$-th nodes inside the same sphere. $\alpha_{j,k}$ is the local attention score between node $p_j$ and the other nodes inside the same sphere. $\beta_m$ is the global attention score from the global KNN graph in the $m$-th iterations.

As shown in Figure~\ref{fig:2} (a), we construct a complete graph for all nodes within a voxel sphere to learn the features constrained by each other. In order to allow each point to attend on every other point and make coefficients easily comparable across different points, we normalize them across all choices using the softmax function, so the local attention score $\alpha_{j,k}$ is calculated by:
\begin{small}
\begin{equation}
\label{eqn:01}
\alpha_{j,k} = softmax_j(f_j,f_{j,k}) = \frac{\exp(f^T_j \cdot f_{j,k})}{\sum_{k\in \sqcup(p_j)}\exp(f^T_j \cdot f_{j,k})}
\end{equation}
\end{small}

\textbf{Global attention layer.} By constructing the local complete graph, the aggregated features can only describe the local feature and do not integrate with the global information. So we design the global attention layer to learn the global feature of each spherical voxel and offer a feature factor aligned to each node.

For the points within each $b_i$ in $N$ 3D spherical voxels $B=\{b_1, b_2, ..., b_N\}$, we calculate the physical centers of all voxels which denoted as $\{c_i\}_{i=1,...,N}$. Each center is learned by a 3-layer MLP to get the initial global feature $F_g=\{f_{g,1}, f_{g,2},...,f_{g,N}\}$. As Figure~\ref{fig:2} (b) shows, we construct a KNN graph for the $N$ voxel sphere. For each node $f_{g,i}$, the attention score between node $f_{g,i}$ and its $l$-th neighbor is calculated as follows:
\begin{equation}
\label{eqn:05}
\beta_m = \frac{f^T_{g,i} \cdot f_{g,i,l}}{\sum_{l\in \mho(f_{g,i})}f^T_{g,i} \cdot f_{g,i,l}}
\end{equation}
where $\mho(f_{g,i})$ denotes the index of the neighbors of node $f_{g,i}$. $m$ is the number of the point attention layers. Eq.~\ref{eqn:05} can be regarded as a weighted summation of the K neighbor nodes around a node, which guarantees the permutation invariance to the nodes' order.


\textbf{Voxel-graph features representation.} The point attention operation on each spherical voxel can combine the parameter factor from both local and global, each of which is inserted with a 2-layer MLP with a nonlinear activation to transform each updated feature $f^{'}_j$. By stacking multiple point attention layers, both local aggregated feature and global point-wise feature can be learned. We then apply maxpool on the aggregated feature to obtain the final feature vector. To process all the spherical voxel, we obtain a set of voxel sphere features, each of which corresponds to the spatial coordinates of the voxels and is taken as input of the sparse-to-dense regression module.
\begin{figure}[htb]
  \centering
  \includegraphics[width=1\linewidth]{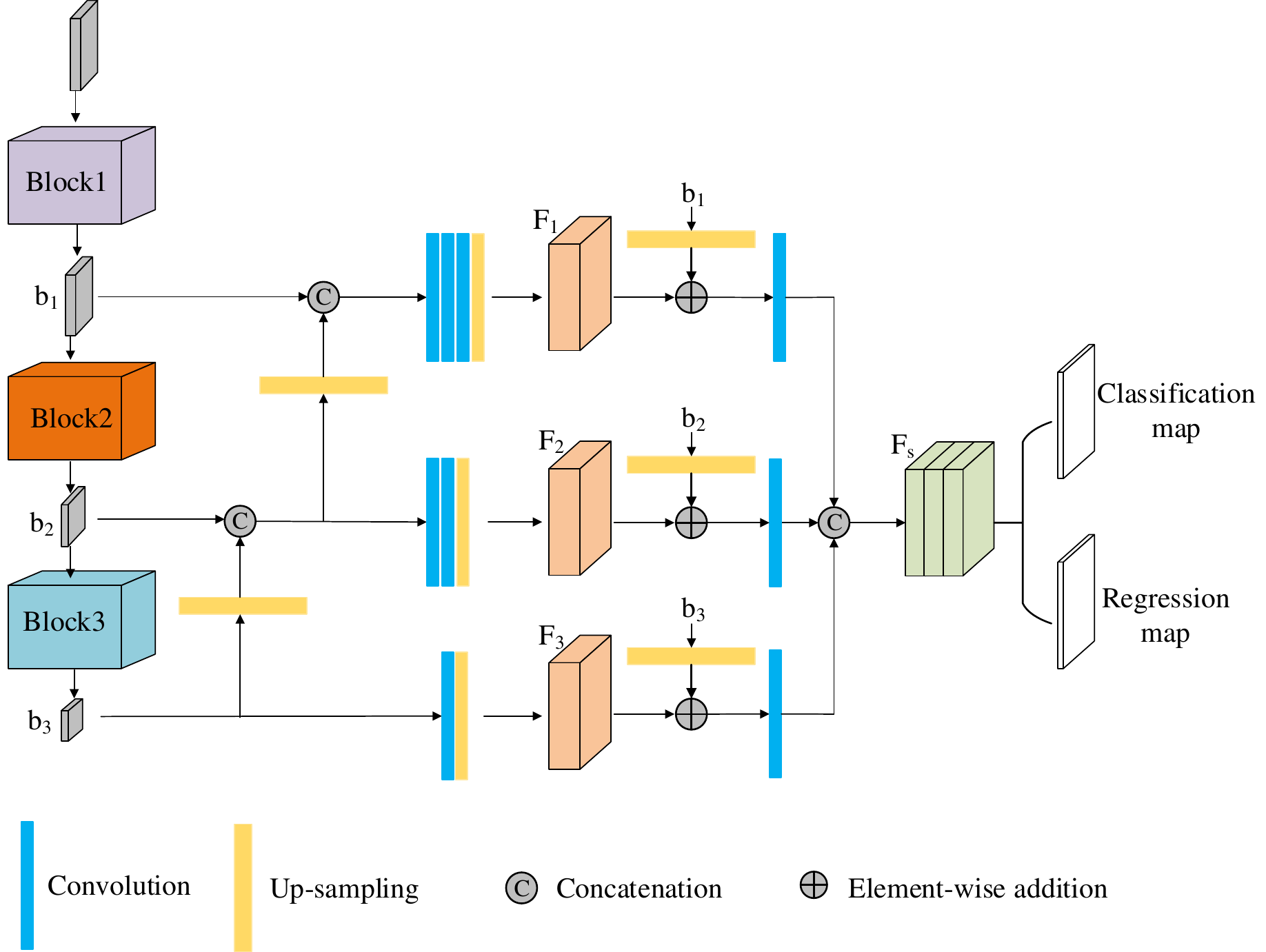}
  \caption{The architecture of the sparse-to-dense regression module. Features from the voxel-graph network are processed by series of region proposal extraction operations to generate the final classification and regression maps.}
  \label{fig:3}
\end{figure}
\subsection{Sparse-to-dense regression}
\label{sec:sim4}

For each 3D bounding box in 3D space, the predicted box information is represented as $(x,y,z,l,w,h,\theta)$, where $(x,y,z)$ is the center coordinate of the bounding box, $(l,w,h)$ is the size information alongside length, width and height respectively, and $\theta$ is the heading angle. Feature map from the voxel-graph network is processed by region proposal regression module. The architecture of the specified sparse-to-dense regression(SDR) module is illustrated in Figure~\ref{fig:3}.

SDR module first apply three similar blocks as \cite{zhou2018voxelnet,lang2019pointpillars} to generate smaller the spatial resolution from top to down. Each block consist of series of Conv$2$D$(f_{in},f_{out},k,s,p)$ layers, followed by BatchNorm and a ReLU, where $f_{in}$ and $f_{out}$ are the number of input and output channels, $k,s,p$ represent the kernel size, stride size and padding size respectively. The stride size is set to 2 for the first layer of each block to downsample the feature map by half, followed by sequence of convolutions with stride 1. And the output of the three blocks is denoted as $b_1$, $b_2$, $b_3$ respectively.

In order to combine high-resolution features with large receptive fields and low-resolution features with small receptive fields, we concat the output of the second and third modules $b_2$, $b_3$ with the output of the first and second modules $b_1$, $b_2$ after upsampling. In this way, the dense feature range of the lower level can be well combined with the sparse feature range of the higher level. Then a series of convolution operations with an upsampling layer are performed in parallel on three scale channels to generate three feature maps with the same scale size, which are denoted as $F_1$, $F_2$, $F_3$.

In addition, we consider that the features output of $F_1$, $F_2$, $F_3$ are more densely fit to our final goal than the original three modules. Therefore, in order to combine the original sparse feature map and the series of processed dense feature maps, we combine the original output $b_1$, $b_2$, $b_3$ after upsampling and $F_1$, $F_2$, $F_3$ by element-wise addition. The final output $F_s$ is obtained by concatenating the fused feature maps after a $3\times3$ convolution layer. And $F_s$ is taken as input to perform category classification and 3D bounding box regression.
\subsection{Loss function}
We use a multi-task loss to train our network. Each prior anchor and ground truth bounding box are parameterized as $(x_a,y_a,z_a,l_a,w_a,h_a,\theta_a)$ and $(x_{gt},y_{gt},z_{gt},l_{gt},w_{gt},h_{gt},\theta_{gt})$ respectively. The regression residuals between anchors and ground truth are computed as:
\begin{equation}
\begin{aligned}
&\Delta x = \frac{x_{gt}-x_a}{d_a}, \Delta y = \frac{y_{gt}-y_a}{d_a}, \Delta z = \frac{z_{gt}-z_a}{h_a} \\
&\Delta w = \log(\frac{w_{gt}}{w_a}), \Delta l = \log(\frac{l_{gt}}{l_a}), \Delta h = \log(\frac{h_{gt}}{h_a})\\
&\Delta\theta = \sin (\theta_{gt} - \theta_a)
\end{aligned}
\end{equation}
where $d_a=\sqrt{(w_a)^2+(l_a)^2}$. And we use Smooth L1 loss\cite{7410526} as our 3D bounding box regression loss $L_{reg}$.

For the object classification loss, we apply the classification binary cross entropy loss.
\begin{small}
\begin{equation}
\label{eqn:01}
L_{cls}\!=\!\gamma_1\frac{1}{N_{pos}}\sum_{i}L_{cls}(p_i^{pos},1)\!+\!\gamma_2 \frac{1}{N_{neg}}\sum_{i}L_{cls}(p_i^{neg},0).
\end{equation}
\end{small}
where $N_{pos}$ and $N_{neg}$ are the number of the positive and negative anchors. $p_i^{pos}$ and $p_i^{neg}$ are the softmax output for positive and negative anchors respectively. $\gamma_1$ and $\gamma_2$ are positive constants to balance the different anchors, which are set to 1.5 and 1 respectively in our practice.

Our total loss is composed of two parts, the classification loss $L_{cls}$ and the bounding box regression loss $L_{reg}$ as:
\begin{small}
\begin{equation}
\label{eqn:01}
L_{total}=\alpha L_{cls}+\beta \frac{1}{N_{pos}}\sum_{t\in\{x,y,z,l,w,h,\theta\}}L_{seg}(\Delta t^{*},\Delta t).
\end{equation}
\end{small}
where $\Delta t^{*}$ and $\Delta t$ are the predicated residual and the regression target respectively. Weighting parameters $\alpha$ and $\beta$ are used to balance the relative importance of different parts, and their values are set to 1 and 2 respectively.

\section{Experiments}
\textbf{KITTI.} We first evaluate our method on the widely used KITTI 3D object detection benchmark~\cite{geiger2012we}. It includes 7481 training samples and 7518 test samples with three categories: car, pedestrian and cyclist. For each category, detection results are evaluated based on three levels of difficulty: easy, moderate and hard. Furthermore, we divide the training data into a training set (3712 images and point clouds) and a validation set (3769 images and point clouds) at a ratio of about 1: 1 (Ablation studies are conducted on this split). We train our model on train split and compare our results with state-of-the-art methods on both val split and test split. For evaluation, the average precision (AP) metric is to compare with different methods and the 3D IoU of car, cyclist, and pedestrian are 0.7, 0.5, and 0.5 respectively.

\begin{table*}[ht]
\centering
\scalebox{0.8}
{
\begin{tabular}{ccccccccccc} \toprule
            \multirow{2}{*}{Method} & \multirow{2}{*}{Modality} & \multicolumn{3}{c}{$AP_{car}(\%)$} & \multicolumn{3}{c}{$AP_{pedestrian}(\%)$} & \multicolumn{3}{c}{$AP_{cyclist}(\%)$} \\
            \cmidrule(r){3-5} \cmidrule(r){6-8} \cmidrule(r){9-11}
                  &    & Easy & Moderate & Hard & Easy & Moderate & Hard & Easy & Moderate & Hard\\ \hline
MV3D\cite{chen2017multi} & R+L  &71.09 & 62.35  & 55.12 & -  & - & -  & - & - & -\\
F-Pointnet\cite{qi2018frustum} & R+L  &81.20 & 70.39  & 62.19 & 51.21 & 44.89 & 40.23  & 71.96 & 56.77 & 50.39\\
AVOD-FPN\cite{ku2018joint} & R+L  & 81.94 & 71.88  & 66.38 & 50.80 & 42.81 & 40.88  & 64.00 & 52.18 & 46.61\\
F-ConvNet\cite{wang2019frustum} & R+L  &85.88 &76.51 & 68.08 & 52.37 & {\bf45.61} & 41.49 & {\bf79.58} & {\bf64.68} & 57.03\\
MMF\cite{liang2019multi} & R+L  &86.81 & 76.75  & 68.41 & -  & - & - & - & - & -\\\hline
Voxelnet\cite{zhou2018voxelnet} &   L  &77.47 & 65.11  & 57.73 & 39.48 & 33.69 & 31.51  & 61.22 & 48.36 & 44.37\\
SECOND\cite{yan2018second} &   L  & 83.13 & 73.66  & 66.20 & 51.07 & 42.56 & 37.29  & 70.51 & 53.85 & 46.90\\
PointPillars\cite{lang2019pointpillars} &   L  &79.05 & 74.99  & 68.30 & 52.08 & 43.43 & 41.49 & 75.78 & 59.07 & 52.92\\
PointRCNN\cite{shi2019pointrcnn} &   L &85.94 & 75.76  & 68.32 & 49.43 & 41.78 & 38.63 & 73.93 & 59.60 & 53.59\\
STD\cite{yang2019std} &   L  & 86.61 & 77.63  & 76.06 & {\bf53.08} & 44.24 & {\bf41.97} & 78.89 & 62.53 & 55.77\\
3DSSD\cite{yang20203dssd} & L & 88.36 & 79.57 & 74.55 & - & - & - & - & - & - \\
SA-SSD\cite{he2020structure} & L & 88.75 & 79.79 & 74.16  & - & - & - & - & - & - \\
PV-RCNN \cite{shi2020pv} &  L   &  {\bf90.25} & {\bf81.43} & {\bf76.82} & - & - & - & 78.60 & 63.71 & {\bf57.65}  \\
Point-GNN\cite{Point-GNN} & L &88.33 & 79.47 & 72.29 & 51.92 & 43.77 & 40.14 & 78.60 & 63.48 & 57.08\\\hline
SVGA-Net(ours) &  L & 87.33 & 80.47 & 75.91 & 48.48  & 40.39 & 37.92 & 78.58 & 62.28 & 54.88\\
            \bottomrule
        \end{tabular}}
\caption{Performance comparison on KITTI 3D object detection for car, pedestrian and cyclists.The evaluation metrics is the average precision (AP) on the official test set. 'R' denotes RGB images input and 'L' denotes Lidar point clouds input.}
\label{tab:one}
\end{table*}
\subsection{Training}
\textbf{Network Architecture.} As shown in Figure~\ref{fig:1}, in the local point-wise feature and global attention layer, the point sets are first processed by 3-layer MLP and the sizes are all (64, 128, 128). In the local point attention layer, we stack $n=3$ local point-attention graph to aggregate the features, each followed by a 2-layer MLP. And the sizes of the three MLPs are (128, 128), (128, 256) and (512, 1024) respectively. Following~\cite{ku2018joint,zhou2018voxelnet,yang2019std}, we train two networks, one for cars and another for both pedestrians and cyclists.

For cars, we sample $N=1024$ to form the initial point sets. To construct the local complete graph, we choose $r=1.8 m$. For anchors, an anchor is considered as positive if it has the highest IoU with a ground truth or its IoU score is over 0.6. An anchor is considered as negative if the IoU with all ground truth boxes is less than 0.45. To reduce redundancy, we apply IoU threshold of 0.7 for NMS. For cyclist and pedestrian, the number of the initial point sets is $n=512$. We set $r=0.8$ to construct the local graph. The anchor is considered as positive if its highest IoU score with a ground truth box or an IoU score is over than 0.5. And an anchor is considered as negative if its IoU score with ground truth box is less than 0.35. The IoU threshold of NMS is set to 0.6.

The network is trained in an end-to-end manner on GTX 1080 GPU. The ADAM optimizer~\cite{kingma2014adam} is employed to train our network and its initial learning rate is 0.001 for the first 140 epoches and is decayed by 10 times in every 20 epoches. We train our network for 200 epoches with a batch size of 16 on 4 GPU cards. Furthermore, we also apply data augmentation as \cite{lang2019pointpillars,zhou2018voxelnet} do to prevent overfitting.
\subsection{Comparing with state-of-the-art methods }
\textbf{Performance on KITTI test dataset.} We evaluate our method on the 3D detection benchmark benchmark of the KITTI test server. As shown in Table~\ref{tab:one}, we compare our results with state-of-the-art RGB+Lidar and Lidar only methods for the 3D object detection and the bird's view detection task. Our proposed method outperforms the most effective RGB+Lidar methods MMF\cite{liang2019multi} by (0.52\%, 3.72\%, 7.50\%) for car category on three difficulty levels of 3D detection.

Compared with the Lidar-based methods, our SVGA-Net can still show decent performance on the three categories. In particular, we achieve decent results compared to Point-GNN\cite{Point-GNN} using the same graph representation method but using graph neural network in the detection of the three categories. We believe that this may benefit from our construction of local and global graphs to better capture the feature information of point clouds. The slight inferiority in the two detection tasks may be due to the fact that the local graph cannot be constructed for objects with occlusion ratio exceeding 80\%.

\begin{table}[ht]
\centering
\scalebox{0.7}{
\begin{tabular}{ccccc} \toprule
            \multirow{2}{*}{Method} & \multirow{2}{*}{Modality}  & \multicolumn{3}{c}{$AP_{car}(\%)$} \\
            \cmidrule(r){3-5}
                  &    & Easy & Moderate & Hard \\ \hline
MV3D~\cite{chen2017multi} & R+L & 71.29 & 62.68  & 56.56 \\
F-Pointnet~\cite{qi2018frustum} & R+L & 83.76 & 70.92  & 63.65 \\
AVOD-FPN~\cite{ku2018joint} & R+L & 84.41 & 74.44  & 68.65 \\
F-ConvNet\cite{wang2019frustum} & R+L & 89.02 & 78.80 & 77.09 \\\hline
Voxelnet~\cite{zhou2018voxelnet} &   L  & 81.97 & 65.46  & 62.85 \\
SECOND~\cite{yan2018second} &   L & 87.43 & 76.48  & 69.10 \\
PointRCNN~\cite{shi2019pointrcnn} &   L    & 88.88 & 78.63  & 77.38 \\
Fast PointRCNN~\cite{chen2019fast} &   L    & 89.12 & 79.00  & 77.48 \\
STD\cite{yang2019std} &   L & 89.70 & 79.80 &{\bf79.30} \\
SA-SSD\cite{he2020structure} & L & 90.15 & 79.91 & 78.78\\
3DSSD\cite{yang20203dssd} & L & 89.71 & 79.45 & 78.67\\
Point-GNN\cite{Point-GNN} & L & 87.89 & 78.34 & 77.38\\\hline
SVGA-Net(ours) &  L  & {\bf90.59} & {\bf80.23}  & 79.15 \\
            \bottomrule
        \end{tabular}
        }
\caption{Performance comparison on KITTI 3D object detection val set for car class.}
\label{tab:3}
\end{table}

\begin{table}[ht]
\centering
\scalebox{0.7}{
\begin{tabular}{ccccc} \toprule
            \multirow{2}{*}{Method} & \multirow{2}{*}{Modality}  & \multicolumn{3}{c}{$AP_{car}(\%)$} \\
            \cmidrule(r){3-5}
                  &    & Easy & Moderate & Hard \\ \hline
MV3D~\cite{chen2017multi} & R+L & 86.55 & 78.10  & 76.67 \\
F-Pointnet~\cite{qi2018frustum} & R+L & 88.16 & 84.02  & 76.44 \\
F-ConvNet\cite{wang2019frustum} & R+L & 90.23 & 88.79 & 86.84\\\hline
Voxelnet~\cite{zhou2018voxelnet} &   L   & 89.60 & 84.81  & 78.57 \\
SECOND~\cite{yan2018second} &   L   & 89.96 & 87.07  & 79.66 \\
Fast PointRCNN~\cite{chen2019fast} &   L   & 90.12 & 88.10  & 86.24 \\
STD\cite{yang2019std} &   L & {\bf90.50} & 88.50 & 88.10 \\
Point-GNN\cite{Point-GNN} & L & 89.82 & 88.31 & 87.16\\\hline
SVGA-Net(ours) &  L  & 90.27 & {\bf89.16}  & {\bf88.11} \\
            \bottomrule
        \end{tabular}
        }
\caption{Performance comparison on KITTI bird's eye view detection val set for car class.}
\label{tab:4}
\end{table}

\textbf{Performance on KITTI validation dataset.} For the most important car category, we also report the performance of our method on KITTI val split and the results are shown in Table~\ref{tab:3} and Table~\ref{tab:4}. For car, our proposed method achieves better or comparable results than state-of-the-art methods on three difficulty levels which illustrate the superiority of our method.

\begin{figure*}[ht]
\centering
  \includegraphics[width=17cm]{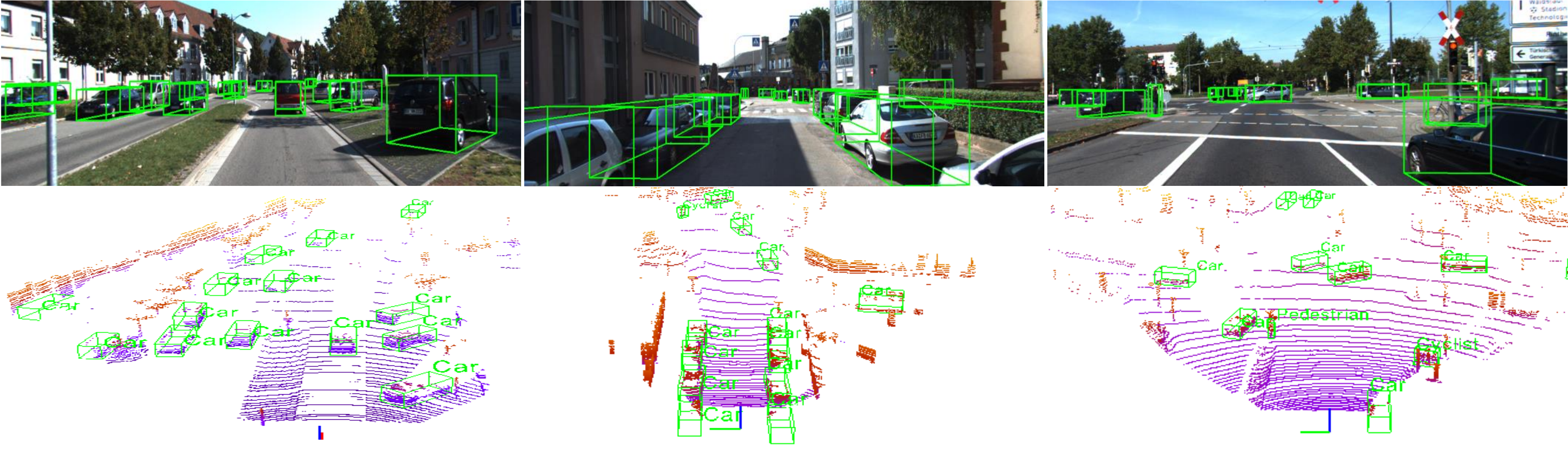}
  \caption{Qualitative 3D detection results of SVGA-Net on the KITTI test set. The detected objects are shown with green 3D bounding boxes and the relative labels. The upper row in each image is the 3D object detection result projected onto the RGB image and the bottom is the result in the corresponding point clouds.}
  \label{fig:6}
\end{figure*}
\subsection{Qualitative results}
As shown in Figure~\ref{fig:6}, we illustrate some qualitative predicted bounding results of our proposed SVGA-Net on the test split on KITTI dataset. For better visualization, we project the 3D bounding boxes into RGB images and BEV in point clouds. From the figures we could see that our proposed network could estimate accurate 3D bounding boxes in different scenes. Surprisingly, SVGA-Net can still produce accurate 3D bounding boxes even under poor lighting conditions and severe occlusion.

\subsection{Ablation studies}
In this section, we conduct series of extensive ablation studies on the validation split of KITTI to illustrate the role of each module in improving the final result and our parameter selection. All ablation studies are implemented on the car class which contains the largest amount of training examples. The evaluation metric is the average precision (AP \%) on the val set.

\textbf{Effect of different design choice.} In the local point attention layer, we stack several local complete layers to extract aggregated features. In order to show the impact of the number of the point attention layer, we train our network with $n$ varying from 1 to 4. As shown in Table~\ref{tab:5}, when the local feature information is transmitted on the 1st to 3rd layers, the detection accuracy is continuously improved because the features are continuously aggregated to the object itself. When $n$ increases to 4, the detection accuracy decreases slightly, and we believe that the network should be over-learning.

Furthermore, we study the importance of the global attention layer in improving the detection accuracy. As shown in Table~\ref{tab:5}, the AP values on both detection tasks are greatly reduced when we remove this module from the network, which proves the importance of this design in providing global feature information for each point.

\begin{table}[ht]
\centering
\scalebox{0.7}{
\begin{tabular}{cccccccc} \toprule
     &  & \multicolumn{3}{c}{$3D AP_{car}(\%)$} & \multicolumn{3}{c}{$BEV AP_{car}(\%)$} \\
            \cmidrule(r){3-5} \cmidrule(r){6-8}
                    &   & Easy & Moderate & Hard & Easy & Moderate & Hard\\\hline
  \multirow{4}{*}{n}&  1  &  86.77 & 75.37 & 74.19 & 87.54 & 86.11 & 83.72\\
                    &  2  &  88.86 & 78.81 & 78.03 & 89.04 & 88.44 & 87.05\\
                    &  3  &  90.59 & 80.23 & 79.15 & 90.27 & 89.16 & 88.11  \\
                    &  4  &  89.62 & 79.26 & 77.58 & 89.72 & 88.51 & 87.17\\\hline
  \multirow{2}{*}{w/o.} &  o.   &  88.42 & 78.11 & 76.54 & 89.71 & 87.45 & 84.33\\
                        &  w.   &  90.59 & 80.23 & 79.15 & 90.27 & 89.16 & 88.11 \\\hline
                        &  SR   &  87.53 & 77.81 & 76.22 & 86.95 & 86.62 & 85.04 \\
                        &  DR   &  88.39 & 78.44 & 76.56 & 87.91 & 86.82 & 86.73 \\
                        &  SDR  &  90.59 & 80.23 & 79.15 & 90.27 & 89.16 & 88.11 \\\hline
  \multirow{5}{*}{k} &  1  &  76.37 & 69.15 & 68.47 & 82.11 & 80.27 & 79.58\\
                      &  2  &  84.53 & 75.61 & 71.92 & 86.23 & 85.65 & 83.66\\
                      &  3  &  90.59 & 80.23 & 79.15 & 90.27 & 89.16 & 88.11  \\
                      &  4  &  88.91 & 79.22 & 77.86 & 88.07 & 87.88 & 87.08\\
                      &  5  &  86.58 & 76.82 & 75.43 & 85.29 & 84.38 & 83.47\\
            \bottomrule
        \end{tabular}
        }
\caption{Performance comparison with different design choice. n is the number of point-attention layers. 'w/o.' denotes whether to keep the global attention layer. SDR denotes the sparse-to-dense regression.}
\label{tab:5}
\end{table}
In the middle three rows of Table~\ref{tab:5}, we aim to explore the effect of different design in the spare-to-dense regression module. SR is to remove the concatenation of $b_1,b_2$ with the upsampled $b_2,b_3$ and DR is to remove the addition of $b_i$ with $F_i$. Results show that only the design of sparse-to-dense regression ranks the first in improving detection accuracy.

When constructing the KNN graph, the number "3" in our implementation is chosen after series of experiments on val set, as shown in the last five rows in Table ~\ref{tab:5}. When $K$ increases from 1 to 3, the AP value has a significant increase, but when it continues to increase, the AP value does decrease.

\textbf{Running time.} Our network is written in Python and implemented in Pytorch for GPU computation. The average inference time for one sample is 62 ms, including 14.5\%(9 ms) for data reading and pre-processing, 66.1\%(41 ms) for local and global features aggregation and 19.4\%(12 ms) for final boxes detection.
\section{Conclusions}
In this paper, we propose a novel sparse voxel-graph attention network(SVGA-Net) for 3D Object Detection from raw Point Clouds. We introduce graph representation to process point clouds. By constructing a local complete graph in the divided spherical voxel space, we can get a better local representation of the point feature, and the information between the point and its neighborhood can be fused. By constructing a global graph, we can better supervise and learn the features of points. In addition, the sparse-to-dense regression module can also fuse feature maps at different scales. Experiments have demonstrated the efficiency of the design choice in our network. Future work will extend SVGA-Net to combine RGB images to further improve detection accuracy.
\section{Acknowledgments}
This work is supported by a grant from the National Natural
Science Foundation of China (No.61872068),
by a grant from Science \& Technology Department of Sichuan
Province of China (No.2020YFG0037, 2020YFG0287,2021YFG0366).

{\small
\bibliographystyle{ieee_fullname}
\bibliography{egbib}
}

\end{document}